\pdfoutput=1
\documentclass[11pt]{article}

\usepackage[]{EMNLP2023}
\usepackage{times}
\usepackage{latexsym}
\usepackage{graphicx}
\usepackage{multicol}
\usepackage{multirow}
\usepackage{adjustbox}
\usepackage{comment}
\usepackage{caption}
\usepackage{mathtools}
\usepackage{amsmath}
\usepackage{algorithm}
\usepackage{url}
\usepackage[justification=centering]{caption}
\usepackage{algpseudocode}
\algblock{Input}{EndInput}
\algnotext{EndInput}
\algblock{Paramters}{EndParameter}
\algnotext{EndParameter}

\usepackage[T1]{fontenc}
\usepackage[utf8]{inputenc}

\usepackage{microtype}

\usepackage{inconsolata}
\newcommand{\noteng}[1]{\textcolor{magenta}{\textbf{NG:} #1}}

\newcommand{\ayan}[1]{\textcolor{brown}{\textbf{ayan:} #1}}

\title{Long Dialog Summarization: An Analysis}

\author{$^1$Ankan Mullick \qquad $^2$Ayan Kumar Bhowmick \qquad $^3$ Raghav R \qquad $^4$ Ravi Kokku \\ $^2$ \bf{Prasenjit Dey  \qquad
$^1$Pawan Goyal \qquad $^{1}$Niloy Ganguly}\\ $^1$Department of Computer Science and Engineering, IIT Kharagpur \\$^2$Merlyn Mind Inc.\qquad
 $^3$ Carnegie Mellon University, USA 
\\\texttt{$^1$ankanm@kgpian.iitkgp.ac.in}, \texttt{\{$^2$ayan, $^2$prasenjit, $^2$ravi\}@merlyn.org}, \\\texttt{$^3$rraghavr@cs.cmu.edu},     \texttt{\{$^1$pawang, $^1$niloy\}@cse.iitkgp.ac.in}}

\begin{document}

%\nolinenumbers

\maketitle
\begin{abstract}
%Dialog summarization has become increasingly important in managing and comprehending large-scale conversations across various domains. It is worth noting that the same summarization approach may not always yield the best results. For example, in a shopping-chatbot scenario, the dialog may summarize to learning of user preferences, whereas in the case of a customer call center, the summary may involve the problem attributes that a user specified, and the final resolution provided. Summarization techniques may vary based on specific requirements, such as context-driven summaries or object-oriented summaries and different models excel in different domains. In this paper, we explore current state-of-the-approaches for long dialog summarization in different domains and show that one single model can not perform well across various different domains for different summarization tasks. We also show different metrics can be useful to show the effectiveness of a model in different domains.

Dialog summarization has become increasingly important in managing and comprehending large-scale conversations across various domains. This task presents unique challenges in capturing the key points, context, and nuances of multi-turn long conversations for summarization. It is worth noting that the summarization techniques may vary based on specific requirements such as in a shopping-chatbot scenario, the dialog summary helps to learn user preferences, whereas in the case of a customer call center, the summary may involve the problem attributes that a user specified, and the final resolution provided. This work emphasizes the significance of creating coherent and contextually rich summaries for effective communication in various applications. %, such as customer service, virtual assistants, and collaborative dialog systems. 
%It is worth noting that the same summarization approach may not always yield the best results. For example, in a shopping-chatbot scenario, the dialog may summarize to learning of user preferences, whereas in the case of a customer call center, the summary may involve the problem attributes that a user specified, and the final resolution provided. Summarization techniques may vary based on specific requirements, such as context-driven summaries or object-oriented summaries and different models excel in different domains. 
We explore current state-of-the-art approaches for long dialog summarization in different domains and benchmark metrics based evaluations show that one single model does not perform well across various areas for distinct summarization tasks. %We also show different metrics can be useful to show the effectiveness of a model in different domains.

\end{abstract}

\section{Introduction}

Dialog summarization has emerged as a crucial aspect of managing and understanding large-scale conversations in various contexts such as online tutoring~\cite{jain2023can}, customer service~\cite{liu2019automatic}, patient consultations~\cite{abacha2023overview,joshi2020dr}, and casual chatbot interactions. While the automatic summarization of text remains inherently challenging due to the complexity of determining and preserving the most relevant content, this paper aims to devise goal-oriented summarization and how current state of the art language models can be utilised effectively to achieve superior performance in the domain of large-scale dialog summarization.

\begin{figure}[!htp]
    \centering
    \vspace{-3mm}
    \begin{adjustbox}{width=0.80\linewidth}
\includegraphics[width=\linewidth]{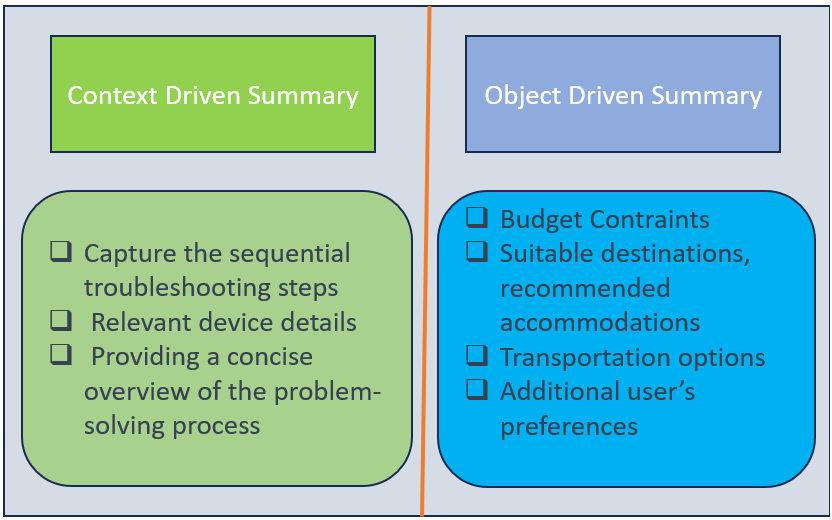}
    \end{adjustbox}
    %\vspace{-1mm}
    \caption{Different Summarization Process}
    \label{fig:intro}
    \vspace{-3mm}
\end{figure}

Summarization will vary according to the requirements - it can be a context driven summary or an object oriented summary as in Fig \ref{fig:intro}. In the former case, in a dialogue session about technical smartphone issues, a context-driven summary would capture the sequential troubleshooting steps and relevant device details, providing a concise overview of the problem-solving process. In the object driven summary, in a travel planning conversation, an objective-driven summary would prioritize the user's goal of obtaining the best itinerary within specified criteria. The summary would highlight suitable destinations, recommended accommodations, transportation options, tailored to the user's preferences and constraints. Thus, the same summarization approach may not lead to the best summaries which will be very useful according to the user's need and leads us to achieve the goal using rubric-driven summarization with the use of custom-trained language models.

%\pg{I feel that these are very good motivations, but the current paper does not deal with most of those. We may need to change the tone accordingly to mention the overall motivation, and what we have been able to do in this survey, coming back to these again in the discussions / conclusions.}

The cornerstone of our approach involves specifying context-driven rubrics in a natural language format, which can guide the fine-tuning of large language models (LLMs) for better goal-oriented summarization performance. As these rubrics address the nuances of different dialog settings and user groups, the fine-tuned models are expected to generate summaries that exhibit a deeper understanding of the conversation, capturing critical information in a coherent and contextually relevant manner. In essence, this paper delves into the foundational objective of enriching the dialog summarization task and create LLM driven solutions that cater to diverse contexts and objectives. By exploring this direction, our work surpasses traditional summarization techniques and offers a compelling groundwork for anyone seeking a deeper comprehension of the intricacies of dialog summarization and its potential applications.

%\ayan{The cornerstone of our approach involves specifying context-driven rubrics in a natural language format, which can guide the fine-tuning of large language models (LLMs) for better goal-oriented summarization performance. As these rubrics address the nuances of different dialog settings and user groups, the fine-tuned models are expected to generate summaries that exhibit a deeper understanding of the conversation, capturing critical information in a coherent and contextually relevant manner.}

%\ayan{In essence, this paper delves into the foundational objective of enriching the dialog summarization task and create LLM driven solutions that cater to diverse contexts and objectives. By exploring this direction, our work surpasses traditional summarization techniques and offers a compelling groundwork for anyone seeking a deeper comprehension of the intricacies of dialog summarization and its potential applications.}
%\noteng{Ayan's part looks much better}

Earlier researchers explore dialog summarization in different directions. %We categorise it into three different components - a) Dialog Summarization  b) Datasets and c) Summarization Metrics.  
%\noindent \textbf{a) Dialog Summarization:}
\citet{vig2022exploring,zhang2022summn,zhu2020hierarchical} provide different approaches for conversation summarization tasks. \citet{xiong2022adapting, pagnoni2022socratic} and FactPegasus~\cite{wan2022factpegasus} study pretraining pipeline for summarizations. Different state-of-the-art models like LongFormer \cite{beltagy2020longformer},  BART~\cite{lewis2019bart}, T5 \cite{raffel2020exploring}, Flan T5 \cite{chung2022scaling} are useful to extract summarized information from dialogs. \citet{leuski2003ineats} present interactive summarization with user control. There exists several benchmarked conversational datasets with summary - QmSum \cite{zhong2021qmsum}, Multiwoz \cite{budzianowski2018multiwoz,eric2020multiwoz}, SParC \cite{yu2019sparc}, SumScreen (Tv series summary) \cite{chen2021summscreen}, SQuALITY \cite{pang2021quality} etc. Some researchers focus on developing evaluation criteria to detect the usefulness of different summarization approaches. ROUGE~\cite{lin2004rouge}, BLEU~\cite{papineni2002bleu}, Meteor~\cite{banerjee2005meteor}, BERTScore\cite{zhang2019bertscore} are state of the art metrics to assess dialog summaries. %\citet{mullick2022evaluation} suggest intent coverage metrics to measure legal document summary. BERTScore\cite{zhang2019bertscore} is proposed to assess text generation. GEM~\cite{gehrmann2021gem} is 

Researchers make impressive advancements on a wide range of downstream applications for summarizations. Since, one approach is not working best across different tasks and different domains, the real-world implications of such advancements are largely unexplored. In this paper, we study utilities of different summarization approaches for extracting compressed contexts from domain specific conversations. Our through extensive experiments on two benchmarked datasets show that different approaches can be useful in different contexts.

\section{Dataset}
We explore two different datasets for the evaluations: 1) \textbf{QmSum \cite{zhong2021qmsum}}: It is a
multi-domain query-based meeting summarization dataset consisting of 1,808 queries and
 232 long meeting transcripts, with different topic of discussions such as software product, academics, and committee meetings. 2) \textbf{SumScreen \cite{chen2021summscreen}}: This summarization dataset comprises  26.9K pairs of TV series
transcripts (`Forever Dreaming' and `TvMegaSite') and human annotated recaps. These TV shows are of different genres like drama, romance, family, etc. Due to the larger size of the datasets, we randomly sample 15\% of the `Forever Dreaming' data from train, dev and test set for our experiment purposes. %In SumScreen, there are two Tv series data - `Forever Dreaming' and `TvMegaSite'. We name them as \textit{SumScreen-FD} and \textit{SumScreen-TM}.

%\noindent Thus in our case, there are three different datasets - \textit{QmSum} and \textit{SumScreen-FD}.% and \textit{SumScreen-TM}. 

%\noteng{are we working on all the three datasets} \ankan{corrected}
\section{Summarization Approaches} \label{model}

We evaluate various state-of-the-art summarization algorithms along with different settings\footnote{Code/Data is available : https://shorturl.at/tyzD8} corresponding to the two datasets discussed above. %Since we experiment with long conversations, we also experiment with various settings to be able to feed the input to the summarization algorithm.

\subsection{Algorithms}
\noindent \textbf{LongFormer \cite{beltagy2020longformer}:} The Longformer (LF) is a transformer-based model designed to handle long documents efficiently %Unlike traditional approaches that require chunking or shortening of long inputs, the Longformer enables seamless processing of extended sequences without the need for complex architectural modifications to integrate information across these chunks. This scalability is achieved through an 
using an attention pattern that effectively combines local and global information, enabling to handle long inputs.% while maintaining linear scalability with the length of the sequence. 

\noindent \textbf{T5 \cite{raffel2020exploring}:} ``Text-to-Text Transfer Transformer'' (T5) model is a  transfer learning technique can be used to generate summary. % by introducing a unified framework that converts all text-based language problems into a text-to-text format. We apply T5 for generating the summary. 
We use T5-base model (770 million parameters).

\noindent \textbf{Flan T5 \cite{chung2022scaling}:} Flan T5 (FT5) is a instruction fine-tuned approach that shows finetuning can improve performance across a range of models, prompting setups, and evaluation tasks. 

\noindent \textbf{BART \cite{lewis2019bart}:} BART is a denoising autoencoder for pretraining sequence-to-sequence models. Pretraining which can generate summary. %has two stages (1) noisy text with an arbitrary noising function, and (2) a sequence-to-sequence model. The later is learned to reconstruct the original text. We apply to generate summaries. 

\noindent \textbf{Chat-GPT~\cite{aydin2023chatgpt,wang2023does}:} We apply ChatGPT (V3.5) to generate summary from dialogs. We apply two different versions of ChatGPT - (P1): where we feed summary and dialog together as text (P2): where we feed the summary as prompt and dialog as text. In (P2), we specify the summary explicitely. 

\subsection{Input Settings}
Since, the conversations exceed the maximum token limit of the models, we follow various settings:

\noindent \textbf{A) Direct:} Apply the model directly on the whole dataset. In this setting, the models automatically truncate the input to the maximum token limit. %\pg{truncating the input to the maximum token limit?} \ankan{edited} like for LongFormer (LF). 
It is shown for the case of LongFormer (LF).

\noindent \textbf{B) Chunk and Summarize:} In this technique, we tokenize the conversation with chunks of maximum token limit and feed individual segments to the models in the following two ways:

i) Feed each conversation segment independently to generate individual summaries of each segment and finally merge the summaries to get the overall summary. In  Table \ref{tab:all_results} these methods are shown as method-name with the chunk size (e.g., LF-8192).

ii)  Generate the summary of the first segment and then feed the obtained summary to the models along with the next conversation segment to generate an updated summary at each step. Continuing with this approach, finally, we get the overall summary from the last conversation segment. This approach is used to preserve the context of a dialog. This is shown as ``type-2'' in the Table \ref{tab:all_results}.

\noindent \textbf{C) Extract then summarize:} In this setting, initially we retrieve sentences %apply state-of-the-art approach \cite{mukherjee2022ectsum} to
from the dialogs (train set) so that the total token size of extracted sentences are within the maximum token limit of the models and then feed it to the summarization frameworks to generate the final summary. The sentences are extracted using the approach similar to \newcite{mukherjee2022ectsum}. %\pg{Clearly mention -- did you use validation set to create training? What extractive summarizer did you use?}
We apply SentenceBERT \cite{reimers2019sentence} to generate the embeddings of the sentences of dialogues and summaries and then calculate the pair-wise cosine similarity. Thereafter, we set a minimum similarity threshold to extract certain dilaog sentences of high similarity ($>$30\%) with the summary sentences. % \noteng{very confusing, rewrite} 
Then, we feed the extracted dialogs to the models to generate summaries (In Table \ref{tab:all_results} these are named as Ex-model like `Ex-T5', `Ex-FT5' etc.). %We apply it on training set and evaluate on unchanged test set.

\section{Experimental Results}
In this section, we evaluate the quality of the final summary generated for a given dialog. For this purpose, we rely on two different approaches, one based on comparing our generated summary using different methods described in Sec.~\ref{model} while the other is based on comparing the coverage of intents and entities in the original dialog with that of the generated summary.
\subsection{Ground truth comparison}
Here we compare the ground truth summary (reference summary) with the generated summary obtained from a given model in terms of how closely the generated summary string matches the ground truth summary string. We measure the extent of similarity between the two summaries with the help of the following standard evaluation metrics: \textbf{1) BLEU score:} BLEU (Bilingual Evaluation Understudy) metric~\cite{papineni2002bleu} measures the precision of n-gram overlap between the generated summary and the reference summary. \textbf{2) ROUGE-Score):} ROUGE (Recall-Oriented Understudy for Gisting Evaluation) is a set of evaluation metrics widely used in summarization tasks~\cite{lin2004rouge}. We use ROUGE-1 (focuses on the unigram or single-word overlap), ROUGE-2 measures the bigram (two-word sequence) and ROUGE-L (considering the longest common subsequence (LCS)). \textbf{3) METEOR:} METEOR (Metric for Evaluation of Translation with Explicit ORdering)~\cite{banerjee2005meteor} evaluates the generated summaries that takes into account the harmonic mean of unigram precision and recall, along with a synonymy matching and sentence structure similarity component. 
\textbf{4) BERTScore \cite{zhang2019bertscore}:} It is a robust technique which computes token similarity using contextual embeddings. It is an useful metric to evaluate summarization models. \textbf{5) Intent-Entity coverage (IEC):}
%Here we extract the set of intents with corresponding entities within a given long dialog conversation. We compare entity set with the corresponding sets obtained using one of the models in Sec.~\ref{model} from the final generated summary for the given dialog conversation. We measure the overlap between the corresponding sets in terms of \emph{Jaccard coefficient}.
To further evaluate the quality of generated summaries in capturing the key information from the original dialog conversation, we propose measuring Intent based entity overlap. These metrics enable a better understanding of how well the generated summaries preserve the primary objectives and capture salient aspects from the conversations. We use \emph{ChatGPT}~\cite{aydin2023chatgpt,wang2023does} in order to extract the set of entities from a given text (dialog conversation or summary), by providing suitable prompts to \emph{ChatGPT}. The overlap in entities is computed as follows:
%\begin{enumerate}
%    \item \textbf{Intent Overlap:} 
    %This metric evaluates the degree to which the summaries preserve the main goals or intentions expressed in the original dialog conversation. 
    To compute the entity Overlap based on \emph{ChatGPT}, we extract the set of entities from both the original dialog conversation and the generated summary using prompts provided to ChatGPT. We then compute the overlap between the two entity sets using \emph{Jaccard similarity} which measures the proportion of common intents in both sets to the combined unique intents.

\begin{table*}[t]
\vspace{-5mm}
\centering
\begin{adjustbox}{width=0.85\linewidth}
\begin{tabular}{|c|c|c|c|c|c|c|c||c|c|c|c|c|c|c|c|c|}
\hline
\multirow{2}{*}{\textbf{Method}} & \multicolumn{7}{c|}{\textbf{QmSum}} & \multicolumn{7}{c|}{\textbf{SumScreen (FD)}}   \\ \cline{2-15} 
                                 & \textbf{Bl} & \textbf{R-1} & \textbf{R-2} & \textbf{R-L}    & \textbf{Mt}& \textbf{BS} & \textbf{IEC}   &  \textbf{Bl} &\textbf{R-1}  & \textbf{R-2} & \textbf{R-L}    & \textbf{Mt}  & \textbf{BS} &  \textbf{IEC}  \\ \hline
LF  & 6.09 & 26.81 & 7.11 & 24.40& 24.76 & 58.96 & 30.81& 0.32	& 15.39 &	1.42	& 12.52 & 3.63  & 47.20& 26.38\\\hline
LF-8192  & 3.58& 23.13 & 5.72& 21.14&18.29 & 56.57& 29.71 &  0.27 &	14.88	& 1.83	& 11.74 &	3.33 & 43.95 & 11.94\\\hline
LF-4096  & 1.74& 22.85&  6.01 & 20.88& 15.45& 55.93& 26.15 & 0.20 & 12.73 &	1.50 &	12.74 &	3.55 & 43.82& 13.71\\\hline
LF-type2  & 0.50 & 	10.87 & 1.50 &	7.01 & 2.4 & 25.95& 23.83&  0.18 & 14.69 & 1.05 & 13.06 & 3.84 &42.91&12.05\\\hline
%Ex-LF  & 72.57 & 87.82 & 83.01 & 87.82 & 77.58 & & & & & & & &&\\\hline
Ex-LF  & 2.49 & 20.84	& 3.75	& 18.37	& 19.08 & 50.34 & 27.15& 0.06 & 17.14	& 1.45 & 15.62	 & 3.16 & 44.79  &23.65\\\hline
T5-4096  & 2.36& 25.00& 7.13& 23.27& 11.55& 59.85 &32.05&   0.35	& 18.57 &	1.48 &	17.62 & 3.82 & 44.99  &31.87\\\hline
T5-2048  & 2.18& 24.80& 6.27& 22.99& 11.61& 55.42 & 29.78 & 0.50 &	19.14	& 1.61 & 18.01 &	5.76 &47.46& 29.36\\\hline
T5-type2  & 2.62 & 21.20& 3.88& 19.80& 19.35& 55.67& 27.42 & 1.15 & 16.84	& 1.22 & 15.19 & 5.07 &47.33&25.86\\\hline
%Ex-T5  & 62.51	& 78.50	& 70.74	& 78.26	& 79.83 & & & & & & & &&\\\hline
Ex-T5  & 7.40 & 24.11 & 7.24 & 22.15 & 22.65 & 54.55 &23.71&  0.99 &	17.25	&1.41 & 15.88 & 13.29&45.41&21.67\\\hline
FT5-2048  & 1.76& 24.00& 5.62& 22.55& 9.04& 55.36 &18.35& 0.47 & 16.86	& 1.41 & 15.54 & 5.51& 46.61&10.24\\\hline
FT5-type2  & 2.62 & 21.20& 3.88& 19.80& 19.35& 57.02 & 21.12& 0.62 & 16.05 & 1.47 & 14.22	& 4.56 & 47.33&17.20\\\hline
%Ex-FT5  & 59.90	& 77.17	& 69.26	& 76.61 &	79.79 & & & & & & & &&\\\hline
Ex-FT5  & 6.84 & 23.88	& 7.29 & 22.41 & 22.23 &54.51 & 28.13& 0.39 & 14.66	& 1.36 & 13.54 & 5.24 &45.51&27.98\\\hline
BART-2048  & 0.90& 24.71& 5.48& 21.06& 13.30&  47.45 &25.74& 0.10 & 19.35 &	2.02 & 17.28 & 8.68& 43.88&23.63\\\hline
BART-type2  & 1.94 & 20.49 & 5.03 & 19.42 & 18.31 &38.20 & 17.32 & 0.39 & 17.03 & 1.37 & 15.54 &	4.29& 44.92& 13.84\\\hline
%Ex-BART  & 68.78	& 82.91	& 76.40	& 82.88	& 84.72  & & & & & & & &&\\\hline
Ex-BART  & 3.10	& 17.29	& 3.54 & 15.94 & 17.36 &47.97  & 21.35 & 0.44	& 22.86	& 1.64 & 22.86 & 12.64 &45.65&18.75\\\hline
ChatGPT  & 0.63	& 15.74	&  2.51 & 13.89 & 9.61 & 54.62 & 25.78&0.40	& 25.04	& 4.34 &18.71 & 18.72 &57.88&28.65\\\hline
ChatGPT P1  & 1.11 & 17.72 & 1.67	& 15.48 &	20.78 & 49.69& 27.91 & 0.35 & 24.82 &	2.77 & 21.83	& 18.74 & 57.10& 30.34\\\hline
ChatGPT P2  & 1.00 & 17.31 & 2.40 & 16.97 & 22.01 &50.98  & 29.12& 2.19 & 22.97 & 3.11 & 	20.17 & 20.32 & 57.33& 31.67\\\hline
\end{tabular}
\end{adjustbox}
\vspace{-2mm}
\caption{Results (in \%) in terms of BLEU (Bl), Rouge (R-1, R-2 and R-L), Meteor (Mt), BERTScore (BS) and Intent-Entity Coverage (IEC) on QmSum and SumScreen (`Forever Dreaming') datasets}
\label{tab:all_results}
\vspace{-5mm}
\end{table*}

\subsection{Results and Discussion}
We present our experimental results in terms of the metrics defined above in Tables~\ref{tab:all_results}. From Table~\ref{tab:all_results}, we observe a diverse range of metric values across the different summary generation models, indicating the varying performance achieved by Longformer, T5, Flan-T5, BART, and ChatGPT in the challenging task of summarizing long dialog conversation. We apply segmentation in the diloags to be fed to the models with different token length as input according to token limits. Different token input lengths are - 8192 for LongFormer (LF); 4096 for for LongFormer (LF) and T5; 2048 for T5, Flan-T5 (FT-5) and BART. Although metric-wise performance varies, some trends are noticeable. For instance, BART demonstrates the lowest BLEU score of 0.90, whereas Longformer attains the highest BLEU score (6.09), ROUGE-1 (26.81) and ROUGE-L (24.40) values. Conversely, T5 registers the lowest values for ROUGE-1 (21.20) and ROUGE-L (19.80) but displays the highest ROUGE-2 score, along with Longformer, at 7.13 and 7.11, respectively. Flan-T5, on the other hand, achieves the lowest ROUGE-2 score of 3.88. In terms of the METEOR metric, Longformer holds the highest score at 24.26, while Flan-T5 records the lowest score of 9.04. For a single approach results also varies for two different datasets. For QmSum, the important conversational sentences which form the summary are mostly located in first and middle segment (77.39\% dialog sentences) of the conversations where as for SumScreen-FD data, 70.1\% of the important conversational sentences which form the summary are from first and middle segment of the dialog. That may explain why LongFormer (LF) without token limit, works well for QmSum but not for SumScreen-FD. %Similarly, there is a drop of scores across all metrics for different approaches.
ChatGPT often takes into account the last segment of the dataset, which may explain why chatGPT produces better outcomes for SumScreen-FD than QmSum.

These contrasting results suggest that each model exhibits specific strengths and weaknesses depending on the aspects of summarization being evaluated. The overall low values for these metrics illustrate the inherent complexity and difficulty of the long dialog summarization task. A possible intuition for such observations is that the ability of the Longformer model to handle long-range contextual information helps it better capture essential content and overall structure. In contrast, the capability of T5 in maintaining local coherence and structure might contribute to its higher ROUGE-2 score, although its lower ROUGE-1 and ROUGE-L scores indicate room for improvement in content coverage. The lower scores of Flan-T5 may be attributed to limitations in its architecture or training data used for the task, indicating a need for further refinement. The varied performance of these models highlights the importance of choosing the most suitable architecture or ensemble strategy based on the desired outcome and evaluation metric for dialog summarization tasks.

%\ayan{AYAN: Need to use the following explanation to relate to ChatGPT's best performance for FD and bad numbers for QMSum}

Compared to models like Longformer, T5, Flan-T5, and BART, ChatGPT may perform better in maintaining contextual relevance and coherence in dialog summarization due to its conversational AI focus. In fact, when we explicitly provide the summary as a promot (P2), ChatGPT performs better in most of the cases. However, it may struggle to capture complete content from long dialogues, as it tends to emphasize recent information. This tendency, caused by limited context window and recency bias, results in summaries focusing on the latter parts of conversations, potentially missing crucial ideas from earlier sections.

\section{Conclusion}

In this paper, we explore how different approaches perform across various long dialog datasets for summarization task. In conclusion, the need for context and objective-driven summarization is evident, as relying on a single approach may not yield the best results. This highlights the significance of rubric-driven summarization techniques and the utilization of custom-trained language models. %This approach ensures the delivery of high-quality summaries tailored to the desired outcomes in various applications.

\section*{Limitations}
Our datasets are not multilingual and multimodal. So, we need to explore how state of the art approaches can be utilized in multlingual and multimodal scenarios - which we aim to do as a part of future work. 

\section*{Ethics Statement}
Our work does not reveal any personal sensitive information and we use publicly available benchmarked datasets and models in different contexts. 

\bibliography{anthology,custom}
\bibliographystyle{acl_natbib}

%\appendix

%\section{Example Appendix}
%\label{sec:appendix}

%This is a section in the appendix.

\end{document}